\documentclass[runningheads]{llncs}
\usepackage{graphicx}
\usepackage{multirow}
\usepackage{booktabs}
\usepackage[misc]{ifsym}
\begin{document}
\title{Accurate Nuclear Segmentation with \\Center Vector Encoding}
%
%
\author{Jiahui Li\and
Zhiqiang Hu\inst{\textrm{\Letter}} \and Shuang Yang}
\authorrunning{J. Li et al.}
%

\institute{SenseTime Research\\
\email{\{lijiahui,huzhiqiang,yangshuang1\}@sensetime.com}}

\maketitle              
\begin{abstract}
Nuclear segmentation is important and frequently demanded for pathology image analysis, yet is also challenging due to nuclear crowdedness and possible occlusion.  In this paper, we present a novel bottom-up method for nuclear segmentation.  The concepts of Center Mask and Center Vector are introduced to better depict the relationship between pixels and nuclear instances.  The instance differentiation process are thus largely simplified and easier to understand. Experiments demonstrate the effectiveness of Center Vector Encoding, where our method outperforms state-of-the-arts by a clear margin.
\end{abstract}
\section{Introduction}
Pathology has long been regarded as the gold standard for medical diagnosis, especially cancer-related diagnosis.  Nuclear segmentation is one of the most important and frequently demanded tasks of pathology image analysis, acting as the building block for nuclear statistical analysis such as size, density, counts, etc., which in turn contribute to cancer grading, therapy planning and outcome prediction.  On the other hand, nuclear segmentation is laborious, tedious, and prone to errors for manual processing, and challenging for automatic methods due to nuclear crowdedness and  possible occlusion.

Nuclear segmentation aims to segment out the nuclear areas from the backgrounds, as well as differentiate nuclear instances.  The task can be well formulated as an instance segmentation problem.  For the instance segmentation problem, there are two main frameworks, namely top-down and bottom-up.  The top-down framework begins by localizing object instances, commonly in the form of bounding boxes, and then performs the segmentation within the bounding box.  Mask R-CNN and its variants \cite{he2017mask,PANet,chen2018masklab} are of the state-of-the-art top-down models.  The bottom-up framework, on the contrary, performs the semantic segmentation on the whole image firstly, and differentiate instances with post-processing based on prior knowledge about the objects \cite{belsare2017breast,NucleiSeg,wang2016automatic,gurcan2009histopathological}.  With the incorporation of prior knowledge, the bottom-up methods are challenging to design, but may perform better than general-purpose top-down counterparts.

In this work we focus on bottom-up methods.  Among the previous works \cite{arbelle2018microscopy,bamford2003automating,fu2014reliable,ho2017nuclei,kumar2017dataset,Li2007,sadanandan2017automated,stegmaier2018cell,su2013cell,wang2017generalizing,xing2016automatic,yin2010cell,zhang2015towards,zhang2015high,zhou2007cell}, Xing et al. \cite{xing2016automatic} proposed a two-class Convolutional Neural Network (CNN) based method for the first-step semantic segmentation that predicts inside and outside mask, and differentiate nuclear instances by carefully-designed distance transform and region growth.  However, the method suffers from under-segmentation, especially for crowded nuclei.   As a follow-up, Kumar et al. \cite{kumar2017dataset} extended the model to a three-class CNN, which predicts boundary mask in addition to inside and outside mask.  They also proposed an anisotropic region growth method for instance differentiation, based on estimated probabilities in the three masks.  Despite the decent improvement compared with Xing et al. \cite{xing2016automatic}, the Kumar et al. \cite{kumar2017dataset} method has at least three limitations:

\begin{enumerate}
\item A simple boundary mask takes limited advantage of overall information.  For example, distance information is not explicitly modeled.
\item The boundary targets during training are sensitive to the nuclear annotations.  Different annotators may deliver slightly different annotations even for the same nucleus, but in this case the boundaries will become totally different.  Since the annotations can never be perfect, the sensitivity will make the model harder to learn.
\item The instance differentiation process is complex.  Multiple thresholds are applied to control the region growth algorithm.  The process behaves more like ad-hoc rules and is hard to tune and understand.
\end{enumerate}

\begin{figure}[!htb]
\includegraphics[width=\textwidth]{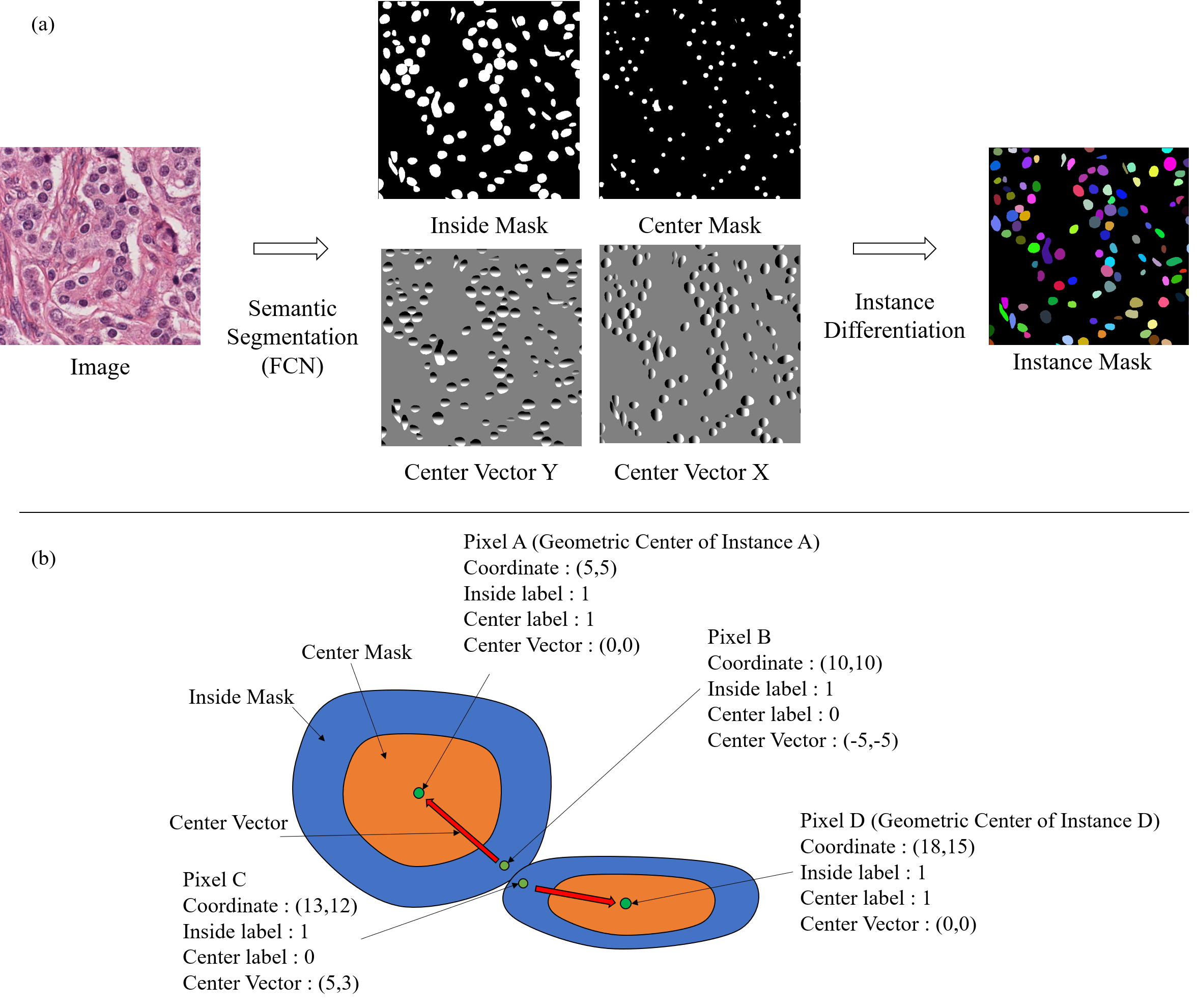}
\caption{(a) Overview of our framework: We apply Fully Convolutional Neural Network (FCN) for semantic segmentation and predicts Inside Mask, Center Mask and Center Vector, which are then utilized in instance differentiation to generate Instance Mask.  (b) Illustration of Center Mask and Center Vector.  Center Mask encodes the center region of nuclei, while Center Vector encodes the relative displacement of each inside-nucleus pixel with respect to the corresponding center.} \label{fig1}
\end{figure}

In this paper, we propose a novel method for nuclear segmentation that overcomes the aformentioned shortcomings.  As shown in Fig. \ref{fig1}, our method begins by Fully Convolutional Neural Network (FCN) \cite{long2015fully,MedicalUnet} for semantic segmentation.  Instead of predicting boundaries, our method estimates the Center Mask, Center Vector, as well as the common Inside Mask.  The Center Mask encodes the center regions of nuclei, the target of which during training is generated by morphological operations and distance thresholding.  The Center Vector, on the other hand, encodes for each inside-nucleus pixel the relative displacement with respect to the centers.  It contains two channels, for horizontal and vertical directions respectively.  During inference, the center region of each nuclear instance is firstly derived by applying connected component analysis to the predicted Center Mask, and then each inside-nucleus pixel (predicted by Inside Mask after some processing) is assigned to a center region according to the predicted Center Vector.  Pixels with Center Vector not pointing to any center region are assigned to the nearest center region.  Finally we refine the Instance Mask, for example filling the holes, to remove the artifacts.

Compared with Kumar et al. \cite{kumar2017dataset}, the concept of Center Vector in our method takes into consideration the distance with respect to the centers, instead of the dichotomous boundary labels, so we utilize more information for the model to separate touching nuclei.  Also, our method is less sensitive to the annotations.  For slighted perturbed annotations, only a tiny fraction of Center Vector targets will be affected, and the model can still learn well based on the majority of correct supervisions.  Lastly, the Center Vector guides the pixels to find its center.  The relationship between pixels and nuclear instances is clear with the concept of Center Vector. The process is straightforward, easier to understand and implement.

We perform extensive experiments on the dataset released by Kumar et al. \cite{kumar2017dataset}.  The quantitative results demonstrate the superiority of our method in performance.  Our method outperforms state-of-the-arts by a clear margin.  We also conduct ablation studies to investigate the benefits gained by introducing Center Vector, both in training phase and in inference phase.  Finally, we show qualitative results to give an insight why the proposed pipeline is better than a boundary-prediction method.

Our contributions are summarized as follows:
\begin{enumerate}
\item We introduce the concepts of Center Mask and Center Vector to better depict the relationship between pixels and nuclear instances.
\item Based on the Center Vector Encoding, we present a pipeline for nuclear segmentation, easy to understand and implement.
\item Our model achieves state-of-the-art performance in the challenging dataset released by Kumar et al. \cite{kumar2017dataset}.  Besides, experiments demonstrate the benefits of Center Vector for better instance differentiation.
\end{enumerate}

\section{Center Vector Encoding}
As shown in Fig. \ref{fig1}(a), the entire pipeline consists of two parts: semantic segmentation and instance differentiation.  We apply Fully Convolutional Neural Network (FCN) \cite{long2015fully} for semantic segmentation and predicts Inside Mask, Center Mask and Center Vector, which are then utilized in instance differentiation to generate Instance Mask.

\subsection{Center Mask and Center Vector in Semantic Segmentation}
Center Mask and Center Vector are the core concepts of our work.  The Center Mask encodes the center regions of nuclei.  The target center region of a nuclear instance during training is calculated as follows: The instance mask for the nuclear instance is firstly processed with morphological erosion, and then we calculate the distance from each pixel to the boundary in the eroded mask, and take as the center region those pixels with distance larger than a threshold.  The operations mainly aim to ensure center regions of  touching nuclei to separate.

The Center Vector, on the other hand, encodes the relative displacement of each inside-nucleus pixel with respect to the corresponding center.  The center $C_i$ for a nuclear instance $i$ is defined as the geometric center of the instance mask, and is denoted as $C_i = \left(\tilde{X}_i, \tilde{Y}_i\right)$.  For each pixel $P_{ij} = \left(X_{ij}, Y_{ij}\right)$ within the instance $i$, the Center Vector $\Delta P_{ij}$ of the pixel is defined as
\begin{equation}
\Delta P_{ij} = P_{ij} - C_i = \left(X_{ij} - \tilde{X_i}, Y_{ij} - \tilde{Y_i}\right)
\end{equation}
For pixels not belonging to any nuclear instance, i.e. the backgrounds, the Center Vector is not defined, and the corresponding loss is ignored during training.  See Fig. \ref{fig1}(b) for an illustration of the concepts of Center Mask and Center Vector.


Compared with boundary encoding \cite{kumar2017dataset}, the Center Vector encodes the distance between pixels and centers instead of the dichotomous boundary labels, and thus contains richer information.  The Center Vector is smooth inside a nuclear instance, making it easier for the model to learn, but changes sharply in the boundary areas, especially in the boundary of touching nuclei, forcing the model to pay more attention to the boundaries.  Moreover, Center Vector is less sensitive to the annotation perturbation.  Even if annotation areas are slightly enlarged or reduced, for example due to different annotation styles by different annotators, only a tiny fraction of Center Vector targets will be affected, and the model can still learn well based on the majority of correct supervision.

Finally, the Inside Mask simply contains all the foreground pixels -- pixels within the overall region of some nuclear instance.  For an input image, the Inside Mask and Center Mask are of the same shape as the image, with 1 channel each in the semantic segmentation output; Center Vector contains 2 channels for horizontal and vertical directions respectively, and is also of the same shape as the input image.  


\subsection{Instance Differentiation}
During inference, the semantic segmentation model outputs three kind of prediction: Inside Mask, Center Mask and Center Vector, from which we will derive the final Instance Mask.  To this end, we firstly perform the connected component analysis to the Inside Mask, and remove regions which have no intersection with Center Mask.  This serves as a kind of false positive suppression.  On the other hand, we also perform the connected component analysis to the Center Mask, and takes each resulting region as the center region of one nuclear instance.  Finally we will assign pixels in the false-positive-suppressed Inside Mask to the center regions: Pixels in the Center Mask are directly assigned to the corresponding center region without considering the Center Vector; For pixels not in the Center Mask but in the Inside Mask, we assign them to the center region their Center Vector points to, or the nearest center region in case that their Center Vector does not point to any valid center region.  Minor refinement are applied afterwards, for example filling the holes, to remove the artifacts.

With the concepts of Center Mask and Center Vector, the instance differentiation process is straightforward and easy to understand.  To summarize, the Center Mask represents the center region of each nuclear instance, while the Center Vector links the pixels to the corresponding center regions.

\section{Experiments}
\subsection{Implementations}
\begin{figure}
\includegraphics[width=\textwidth]{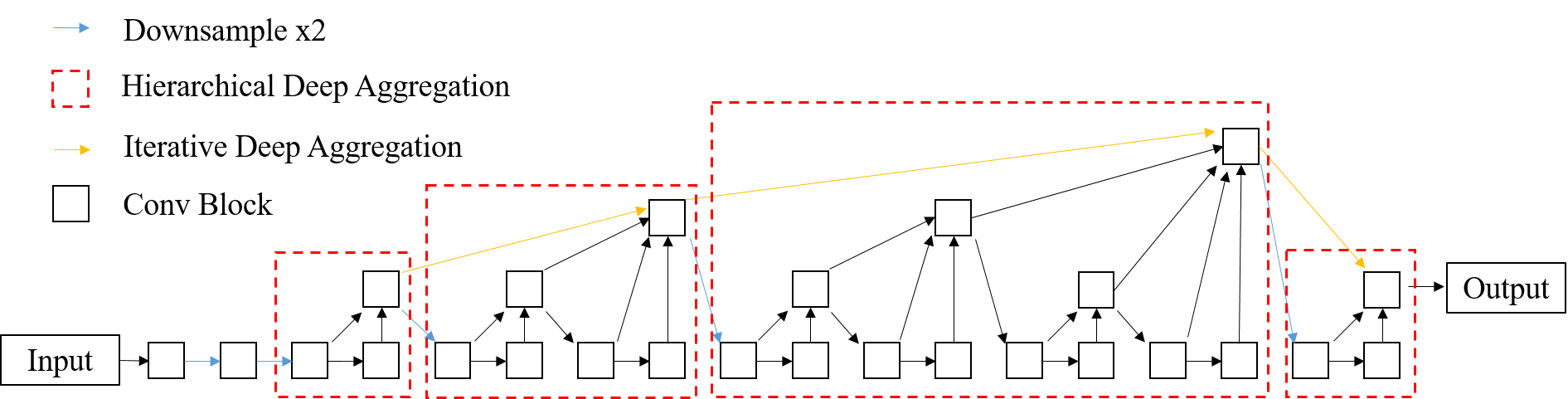}
\caption{Illustration of the Iterative Deep Aggregation and Hierarchical Deep Aggregation introduced by Deep Layer Aggregation.  Iterative Deep Aggregation merges layers iteratively, while Hierarchical Deep Aggregation aggregates layers in a tree-like hierarchical manner.} \label{fig2}
\end{figure}

It is worthy noting that our method does not rely on any particular network architecture for the semantic segmentation model.  For implementation we use Deep Layer Aggregation (DLA) \cite{yu2017deep} for the network architecture in this paper.  DLA extends common network structures with deep aggregations.  The term deep aggregation refers to aggregations that are nonlinear, compositional and going through multiple stages.  DLA introduces two types of deep aggregations: Iterative Deep Aggregation and Hierarchical Deep Aggregation.  Iterative Deep Aggregation merges layers iteratively, while Hierarchical Deep Aggregation aggregates layers in a tree-like hierarchical manner.  With these two types of deep aggregations, the network can better fuse information from multiple layers and scales, and thus achieves better performance in various classification and segmentation problems.  Fig. \ref{fig2} illustrates the Iterative Deep Aggregation and Hierarchical Deep Aggregation introduced by DLA.

For the loss function, we utilize a combination of pixel-wise cross entropy (CE) loss and Intersection-Over-Union (IOU) loss for the optimization of Inside Mask (IM) and Center Mask (CM) estimation.  For Center Vector (CV, with $\mathrm{CV_X\ and \ CV_Y}$ for two directions) with continuous target values, we apply pixel-wise mean square (MS) loss.  Please see (2), (3), and (4) for detailed formulations, where $y_i$ and $p_i$ are targets and predictions for pixel $i$ respectively, i.e., class labels/probabilities for CE and IOU loss but distance targets/estimations for MS loss.  Finally, the total loss is a weighted summation of all the aformentioned losses, as shown in (5), where $\alpha$ , $\beta$ , $\gamma$ are the balancing parameters.

\begin{eqnarray}
l_{\mathrm{CE}} & = & \sum_{S \in \{\mathrm{IM, CM}\}} \sum_{i \in S} y_i \log(p_i) + (1-y_i) \log(1-p_i)\\
l_{\mathrm{IOU}} & = & \sum_{S \in \{\mathrm{IM, CM}\}} 1 - \frac{\sum_{i \in S} y_i p_i}{\sum_{i \in S} y_i + \sum_{i \in S} p_i - \sum_{i \in S} y_i p_i}\\
l_{\mathrm{MS}} & = & \sum_{S \in \{\mathrm{CV_X, CV_Y}\}} \sum_{i \in S} (p_i - y_i)^2\\
l & = & \alpha l_{\mathrm{CE}} + \beta l_{\mathrm{IOU}} + \gamma l_{\mathrm{MS}}
\end{eqnarray}

We use DLA-34 for the model.  The model is implemented in Pytorch 0.4.  We train the model for 2000 epochs. Using SGD as the optimizer, learning rate is initially set as 0.01 and decayed polynomially. Batchsize is 4 and weight decay is 0.0001. In loss function we set $\alpha = 10$, $\beta = 10$, $\gamma = 1$ to balance different losses.

\subsection{Dataset and Evaluation Metric}
We perform extensive experiments on the dataset released by Kumar et al. \cite{kumar2017dataset} to evaluate the proposed method.  The dataset consists of 30 Hematoxylin and Eosin (H\&E) stained images, all of which are of the size $1000\times1000$ pixels.  Nuclear instances are carefully annotated so as to generate the instance mask for each image. The images come from multiple organs: 4 organs (breast, kidney, liver and prostate) have 6 images each, while another 3 organs (bladder, colon and stomach) have 2 images each.  Following the same principle as Kumar et al. \cite{kumar2017dataset}, that is test data must contain images from organs never seen in train data, we sample 4 images from each of 4 majority-class organs (breast, kidney, liver and prostate), a set of 16 images in total for training and put the rest for test.  The multi-organ diversity and zero-shot setting make the problem even more challenging, but closer to clinical practice.

The performance of nuclear segmentation solutions is evaluated based on the metric Aggregated Jaccard Index (AJI) \cite{kumar2017dataset}.  AJI penalizes both segmentation errors and instance differentiation errors, and is a balancing metric for a comprehensive evaluation.  We summarize the AJI calculation steps listed in  \cite{kumar2017dataset} as the equation (6), where $\{G_i\}_{i=1}^{M}$ and $\{P_j\}_{j=1}^{N}$ denote the set of annotated nuclei and predicted nuclei, respectively; $f(i) = \mathrm{argmax}_j |G_i \bigcap P_j| / |G_i \bigcup P_j|$ assigns best-match prediction to each annotated nuclear instance; $[N] = \{1, 2, ..., N\}$ and $f([N]) = \{f(1), f(2), ..., f(N)\}$ are convenient notations.

\begin{eqnarray}
\mathrm{AJI} & = & \frac{\sum_{i=1}^{M}|G_{i}\bigcap P_{f(i)}|}{\sum_{i=1}^{M}|G_{i}\bigcup P_{f(i)}| + \sum_{j\in \left[N\right] \backslash f(\left[M\right])} |P_{j}|}\\
\mathrm{IOU} & = & \frac{|G \bigcap P|}{|G \bigcup P |}\\
\mathrm{Dice} & = & \frac{2 |G \bigcap P|}{|G| + |P|}
\end{eqnarray}

To investigate the error modes, we also apply two metrics originally designed for the semantic segmentation task: Intersection-Over-Union (IOU) and Dice.  Please refer to (7) and (8) for detailed formulations, where $G = \bigcup_{i=1}^{M} G_i$ and $P = \bigcup_{j=1}^N P_j$ are the global annotations and predictions, respectively.  Note that IOU is the upper bound of AJI: We have
\begin{equation}
\sum_{i=1}^{M}|G_{i}\bigcap P_{f(i)}| \le |G \bigcap P|
\end{equation}
and
\begin{equation}
\sum_{i=1}^{M}|G_{i}\bigcup P_{f(i)}| + \sum_{j\in \left[N\right] \backslash f(\left[M\right])} |P_{j}| \ge |G \bigcup P|
\end{equation}
leading to
\begin{equation}
\mathrm{AJI} = \frac{\sum_{i=1}^{M}|G_{i}\bigcap P_{f(i)}|}{\sum_{i=1}^{M}|G_{i}\bigcup P_{f(i)}| + \sum_{j\in \left[N\right] \backslash f(\left[M\right])} |P_{j}|} \le \frac{|G \bigcap P|}{|G \bigcup P |} = \mathrm{IOU}
\end{equation}
Dice is yet the upper bound of IOU.  As a result, the metrics IOU and Dice can be used for distinguishing instance differentiation errors from segmentation errors.

\subsection{Results}
\begin{table}[!t]
\caption{Performance comparison of our method against state-of-the-art methods: CNN2 \cite{xing2016automatic} for two-class CNN (Inside, Outside) and CNN3 \cite{kumar2017dataset} for three-class CNN (Inside, Outside, Boundary).}\label{tab1}
\centering
{\renewcommand{\arraystretch}{1.5}
\begin{tabular}{ccccccc}
\toprule
\multirow{2}*{~~~Organs~~~} & \multicolumn{3}{c}{AJI} & \multicolumn{3}{c}{Dice}\\
\cmidrule(lr){2-4} \cmidrule(lr){5-7}
&~~~CNN2~~~&~~~CNN3~~~&~~~ours~~~&~~~CNN2~~~&~~~CNN3~~~&~~~ours~~~\\
\midrule
Breast& 0.425&0.538&0.601& 0.646&0.718&0.819\\
Liver& 0.371&0.516&0.601& 0.610&0.688&0.797\\
Kidney& 0.407&0.508&0.487& 0.656&0.722&0.737\\
Prostate& 0.228&0.573&0.615& 0.704&0.792&0.792\\
Bladder& 0.319&0.437&0.557& 0.715&0.781&0.815\\
Colon& 0.308&0.521&0.452& 0.664&0.739&0.745\\
Stomach& 0.379&0.446&0.612& 0.855&0.895&0.848\\
\midrule
Overall& 0.348&0.508&\textbf{0.561}& 0.693&0.762&\textbf{0.793}\\
\bottomrule
\end{tabular}
}
\end{table}

\begin{table}[!b]
\caption{Results of ablation studies.  Center Vector is not utilized in V1, only during training in V2, and both during training and inference in V3.  We add results from CNN3 \cite{kumar2017dataset} for comparison.}\label{tab2}
\centering
{\renewcommand{\arraystretch}{1.5}
\begin{tabular}{ccccccccccccc}
\toprule
\multirow{2}*{Organs} & \multicolumn{4}{c}{AJI} & \multicolumn{4}{c}{IOU}& \multicolumn{4}{c}{Dice}\\
\cmidrule(lr){2-5} \cmidrule(lr){6-9} \cmidrule(lr){10-13}
&CNN3&V1&V2&V3&CNN3&V1&V2&V3&CNN3&V1&V2&V3\\
\midrule
Breast&0.538&0.597&0.595&0.601&None&0.699&0.677&0.694&0.718&0.822&0.807&0.819\\
Liver&0.516&0.598&0.573&0.601&None&0.672&0.646&0.663&0.688&0.803&0.784&0.797\\
Kidney&0.508&0.496&0.483&0.487&None&0.606&0.574&0.585&0.722&0.754&0.727&0.737\\
Prostate&0.573&0.600&0.626&0.615&None&0.645&0.648&0.655&0.792&0.784&0.786&0.792\\
Bladder&0.437&0.504&0.544&0.557&None&0.673&0.677&0.693&0.781&0.799&0.803&0.815\\
Colon&0.521&0.441&0.437&0.452&None&0.614&0.560&0.594&0.739&0.763&0.717&0.745\\
Stomach&0.446&0.558&0.593&0.612&None&0.734&0.717&0.736&0.895&0.846&0.835&0.848\\
\midrule
Overall&0.508&0.542&0.550&\textbf{0.561}&None&\textbf{0.664}&0.663&0.663&0.762&\textbf{0.794}&0.791&0.793\\
\bottomrule
\end{tabular}
}
\end{table}

In Table \ref{tab1} we show the performance comparison of our method against state-of-the-art methods: CNN2 \cite{xing2016automatic} for two-class CNN (Inside, Outside) and CNN3 \cite{kumar2017dataset} for three-class CNN (Inside, Outside, Boundary).  We omit IOU scores since they are not reported in the literature.  From the comparison it is clear to see that our method outperforms the state-of-the-arts by a large margin, both with respect to the instance-level AJI metric and with respect to the segmentation-level Dice metric.

We also perform ablation studies to investigate the benefits gained from the design.  We train the semantic segmentation model to predict Inside Mask and Center Mask only, and differentiate instances by Random Walker \cite{RandomWalker} seeded with center regions derived from Center Mask predictions.  We term this setting as V1.  Further for the setting V2, we add Center Vector supervisions during training, but we ignore the Center Vector output during inference and also apply Random Walker \cite{RandomWalker}.  In this case, Center Vector serves as a way to facilitate training.  Finally the setting V3 is the entire method we describe above, where Center Vector is utilized both in training and during inference.

Table \ref{tab2} shows the results of ablation studies.  We add results from CNN3 \cite{kumar2017dataset} for comparison.  The performance gain of V1 from CNN3 \cite{kumar2017dataset} mainly attributes to the better Deep Layer Aggregation (DLA) \cite{yu2017deep} architecture.  Comparing V1, V2 and V3, the AJI score is increasing, which demonstrates the effectiveness of Center Vector, both in training and during inference.  However, the IOU and Dice scores are largely comparable for V1, V2 and V3, showing that Center Vector works mainly from suppressing errors in the instance differentiation step.

\begin{figure}[!b]
\centering
\includegraphics[width=\textwidth]{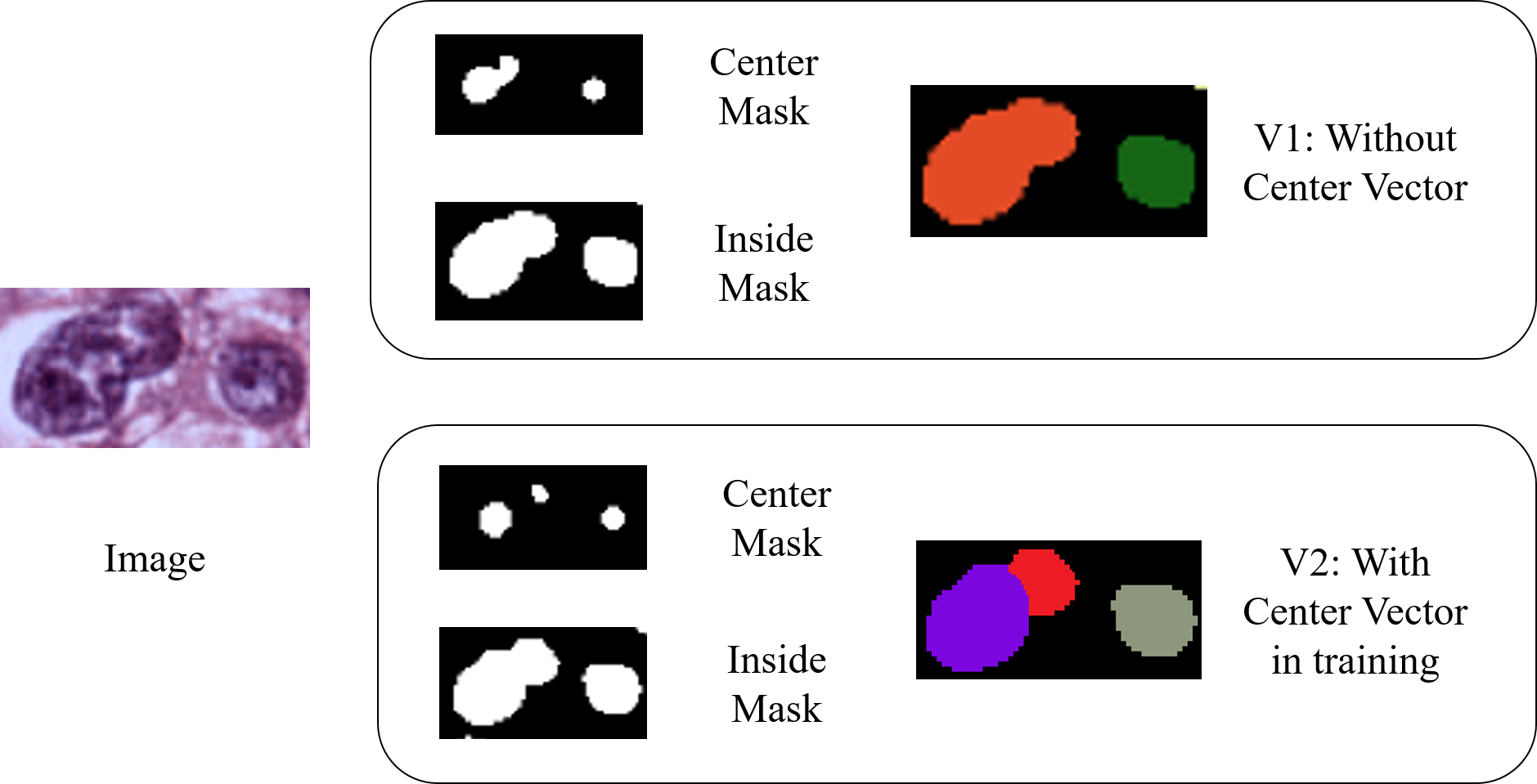}
\caption{Qualitative results from V1 and V2: The V2 model, with Center Vector during training, learns to better separate touching nuclei in the Center Mask.} \label{fig5}
\end{figure}
\begin{figure}[!t]
\centering
\includegraphics[width=0.8\textwidth]{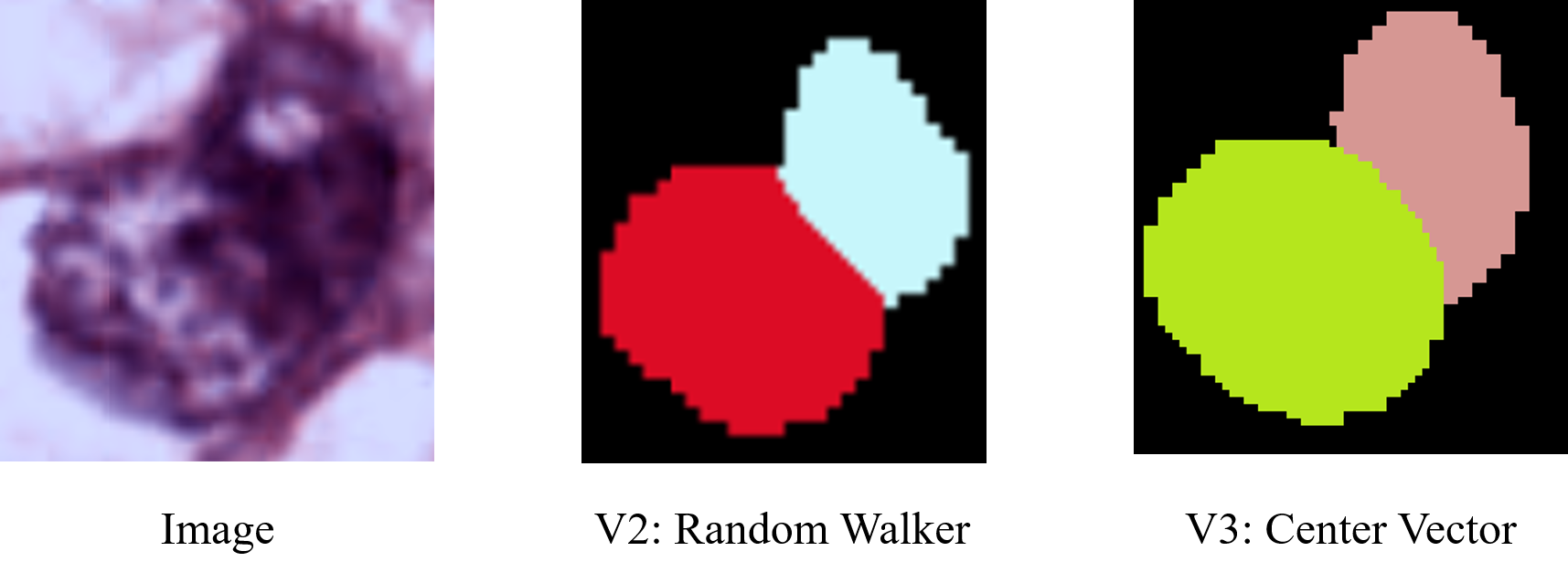}
\caption{Qualitative results from V2 and V3: Random Walker method in V2 tends to separate touching nuclei with a ``straight cut" while Center Vector generates more natural and realistic boundaries.} \label{fig6}
\end{figure}

We also display some qualitative results to give an insight why Center Vector is effective, as shown in Fig. \ref{fig5} and Fig. \ref{fig6}.  Fig. \ref{fig5} compares qualitative results from V1 and V2. The V2 model, with Center Vector during training, learns to better separate touching nuclei in the Center Mask, showing that Center Vector supervision guides the model to concentrate more in the center regions and learn better Center Mask estimation.  On the other hand, Fig. \ref{fig6} compares V2 with V3, where the models differ only in the instance differentiation step.  It can be seen that Random Walker method in V2 tends to separate touching nuclei with a ``straight cut" while Center Vector generates more natural and realistic boundaries.

\section{Conclusion}
We present a novel bottom-up method for nuclear segmentation.  The concepts of Center Mask and Center Vector are introduced to better depict the relationship between pixels and nuclear instances.  Based on the Center Vector Encoding, we develop a pipeline for nuclear segmentation, easy to understand and implement.  Experiments demonstrate the effectiveness of Center Vector Encoding, where our method outperforms state-of-the-arts by a clear margin.


%
%
%
\bibliographystyle{splncs04}

%
%
%
%
%
\end{document}